\documentclass[11pt]{article}

\usepackage[utf8]{inputenc}
\usepackage[T1]{fontenc}
\usepackage{amsmath, amsfonts, amssymb}
\usepackage{graphicx}
\usepackage[colorlinks=true, linkcolor=blue, citecolor=blue, urlcolor=blue]{hyperref}
\usepackage[numbers,sort&compress]{natbib}
\usepackage{geometry}
\usepackage{booktabs}
\usepackage{array}
\usepackage{tabularx}
\usepackage{lmodern}
\usepackage[dvipsnames]{xcolor}
\usepackage{authblk}
\usepackage{setspace}

\begin{document}

\doublespacing

% =============================================================================
% Title and authors
% =============================================================================
\title{Contrast-invariant deep ptychography neural networks}

\author[1]{A. Vong}
\author[1]{S. Henke}
\author[2]{O. Hoidn}
\author[1]{H. Ruth}
\author[1]{J. Deng}
\author[2]{A. Mehta}
\author[3]{D. Shapiro}
\author[3]{A. Hexemer}
\author[1]{N. Schwarz}
% TODO: Add remaining authors

\affil[1]{Argonne National Laboratory, Lemont, Illinois 60439, USA}
\affil[2]{SLAC National Accelerator Laboratory, Menlo Park, California 94025, USA}
\affil[3]{Lawrence Berkeley National Laboratory, 1 Cyclotron Road, Berkeley, California 94720, USA}
% TODO: Add remaining affiliations

\affil[ ]{*Corresponding author: N. Schwarz: nschwarz@anl.gov}

\maketitle

% -----------------------------------------------------------------------------
\begin{abstract}

Ptychography neural networks suffer from scaling inconsistencies when generalizing out of distribution, limiting their real world viability. We address this scaling mismatch using a factorization strategy which decouples the learned object texture from measurement scaling, enabling a single trained network to produce measurement-consistent reconstructions across varying illumination conditions. This requires predicting the learned object in real and imaginary units instead of the canonical amplitude and phase representation. We additionally introduce a synthetic object sampling strategy that minimizes phase distribution mismatch between synthetic training data and experimental targets. These improvements yield up to a 5x reduction in Fourier error over the previous PtychoPINN-torch baseline across 5
experimental datasets spanning multiple beamlines and facilities.

\end{abstract}

% -----------------------------------------------------------------------------
\section{Introduction}
\label{sec:introduction}

Ptychography is a scanning coherent diffractive imaging (CDI) technique that has grown increasingly prominent at light source facilities such as synchrotrons and X-ray free-electron lasers (XFELs), where higher photon fluxes from source upgrades continue to improve achievable spatial resolution. By collecting a series of diffraction patterns from overlapping illumination positions, ptychography overcomes the phase retrieval problem with the assistance of computational algorithms, reconstructing both the measured object and illuminating probe from intensity-only measurements~\cite{faulkner2004movable, thibault2008high}. This technique has enabled advances in fields such as biological imaging~\cite{jiang2010quantitative}, microelectronics metrology~\cite{holler2017high} and nanostructured materials characterization~\cite{michelson2022three}, with spatial resolution surpassing the limits of traditional lens-based optics~\cite{miao2025computational}.

Reconstruction of ptychographic datasets currently relies on iterative phase retrieval algorithms that are computationally expensive and computation time scales poorly with dataset size~\cite{maiden2009improved}. This computational cost has become a critical bottleneck as next-generation light source upgrades and modern photon-counting detectors increase data acquisition rates by several orders of magnitude~\cite{babu2023deep, aps_computing_strategy, schwarz2020enabling}. The mismatch between acquisition and reconstruction throughput is consequential for experiment steering, where rapid qualitative feedback during measurements could accelerate decision-making, impacting data quality and beamtime efficiency.

Feed-forward deep neural networks (DNNs) have been proposed as a solution to this problem: trained models can reconstruct ptychographic data orders of magnitude faster than iterative methods, paving the way for real-time experiment feedback~\cite{lee2025deep, vu2025pid3net, chang2023deep, wu2021complex, gan2024ptychodv, cherukara2020ai, nakahata2024ptychoformer, guan2019ptychonet, cam2025ptychographic, wu2024fourier, wang2024use}. In practice, models may require fine-tuning to generalize to changing measurement conditions such as different samples and photon fluxes, adding processing latency that negates their speed advantage. Recent work has improved object generalization using synthetic data augmentation~\cite{vong2025generalizable} but still produces artifacts such as incorrect amplitude scaling resulting from poor measurement consistency. Here, we define \textit{measurement-consistency} as the reproduction of measured diffraction intensities when the predicted object is propagated through a fully-defined physics forward model.

A leading cause of this generalization failure lies in the object representation and the inherent scale ambiguity of ptychographic reconstruction, as the absolute magnitudes of probe and object cannot be independently determined by the measured intensity. Because the exit wave is defined as a product of object and probe, fixing the probe resolves this ambiguity: the object must then take the correct scale to produce a measurement-consistent diffraction pattern. Recovering this scale requires correcting two degrees of freedom corresponding to the magnitudes of real and imaginary object components. For DNNs using the prevailing amplitude/phase representation, these correction parameters couple nonlinearly in the far-field intensity expression, making it difficult to decouple the individual scaling contributions and formulate a simple optimization objective in intensity space. As a result, learned object textures remain tied to the measurement scales of the training data, preventing tractable rescaling without additional fine-tuning. To our knowledge, no existing strategies combine a neural network representation that factorizes the learned object texture from the measurement-dependent scale, with a principled, training-free procedure to determine the correct scaling for any new dataset.

In this work, we build on our previous unsupervised neural network, PtychoPINN-torch, and present four improvements that collectively address limitations found in previous work:

\begin{enumerate}
    \item \textbf{Real/imaginary decoder output.} We modify the decoder to output normalized rectangular components $(\tilde{a}_o, \tilde{b}_o)$, replacing the previous amplitude/phase representation. This strategy provides a more stable optimization landscape and enables a more principled scaling approach.
    \item \textbf{Dynamic scaling factorization.} We show that with real/imaginary output, the predicted diffraction intensity becomes a quadratic function of two real-valued scaling parameters $(s_1, s_2)$ for real and imaginary predictions, respectively. These scaling constants are optimized via gradient-based methods during test time, ensuring the network-predicted textures are consistent with measured data.
    \item \textbf{Probe-weighted patch stitching.} We replace uniform averaging with $|P|^2$-weighted reassembly, where the probe intensity serves as a spatially varying confidence weight for neural network predictions. This results in large signal to noise improvements in the overall object reconstruction.
    \item \textbf{Synthetic object sampling.} When training with synthetic data, we sample synthetic object components in rectangular form using data-driven correlations rather than uniform amplitude-phase sampling, resolving the phase distribution mismatch identified in our previous work on PtychoPINN-torch.
\end{enumerate}

We demonstrate that these improvements yield quantitative reductions in Fourier error across multiple experimental datasets, alongside qualitative improvements to both amplitude and phase reconstructions. We emphasize that this strategy is universally applicable to unsupervised learning approaches, as the exact DNN architecture can be swapped out for more expressive architecture (e.g. Transformer) with no additional modification to the strategy.

% -----------------------------------------------------------------------------
\section{Methods}
\label{sec:methods}

\subsection{Contrast-invariant neural network decoders}
\label{sec:real_imag_output}

Ptychography neural networks map diffraction images to complex-valued objects, which typically involves an encoder to compress diffraction images to a lower-dimensional latent space, and a decoder which extracts complex-valued images from the latent. We introduce an alternative representation for the decoder output based on rectangular coordinates, i.e. real and imaginary outputs, rather than the canonical amplitude-phase representation favored by ptychography DNNs~\cite{cherukara2020ai, guan2019ptychonet, babu2023deep, gan2024ptychodv, nakahata2024ptychoformer, yamada2024ptychographic}. This formulation allows us to devise a more principled data-driven unit scaling strategy which we describe below. The model incorporating this alternate representation and scaling strategy will be referred to as PtychoPINN-Contrast Invariant (PtychoPINN-CI), versus the amplitude-phase representation (PtychoPINN-torch).

The amplitude-phase decoder typically has a ReLu activation on the amplitude branch (due to unstable losses with more restrictive activation functions) and a hyperbolic tangent on the phase branch, regulating its output to [$-\pi$, $\pi$] \cite{vong2025generalizable}. We replace the amplitude and phase branches to instead output normalized real and imaginary object components, $\tilde{a}_o(\theta)$ and $\tilde{b}_o(\theta)$, where $\theta$ denotes the network weights, each bounded to $[-1, 1]$ by a $\tanh$ activation. The complex object can be reconstructed as
\begin{equation}
    O = a_o + j\, b_o = s_1 \tilde{a}_o(\theta) + j\, s_2 \tilde{b}_o(\theta),
    \label{eq:object_decomp}
\end{equation}
where $s_1$ and $s_2$ are real-valued scalar parameters that absorb the physical amplitude of each rectangular component, converting normalized outputs to amplitude units. This factored representation separates the normalized object texture $(\tilde{a}_o(\theta), \tilde{b}_o(\theta))$, which the network learns, from the dataset-dependent scaling $(s_1, s_2)$, which is optimized independently. The key insight enabling this decoupling is that the predicted intensity becomes a simple quadratic function of $(s_1, s_2)$. Note that the parametrization of the predicted objects using the neural network parameters $\theta$ is assumed implicitly.

Starting from the raw, neural network object texture prediction $O = \tilde{a}_o(\theta) + j\, \tilde{b}_o(\theta)$, we apply the scaling constants $s_1$ and $s_2$ and apply a physics forward model to propagate the exit wave to the far field, giving us a scaled intensity map. We assume a known probe function $P = a_p + j\, b_p$, which remains constant between training and inference.

Substituting Eq.~\eqref{eq:object_decomp} into the exit wave $\psi = P \cdot O$ and expanding in rectangular form yields
\begin{equation}
    \psi = P \cdot O
    = s_1\left(a_p \tilde{a}_o(\theta) + j\, b_p \tilde{a}_o(\theta)\right)
    + s_2\left(-b_p \tilde{b}_o(\theta) + j\, a_p \tilde{b}_o(\theta)\right),
    \label{eq:exit_wave_expansion}
\end{equation}

Defining the Fourier-domain basis components

\begin{align}
    \Psi_a &= \mathcal{F}\left[a_p \tilde{a}_o(\theta) + j\, b_p \tilde{a}_o(\theta)\right], \label{eq:psi_a} \\
    \Psi_b &= \mathcal{F}\left[-b_p \tilde{b}_o(\theta) + j\, a_p \tilde{b}_o(\theta)\right], \label{eq:psi_b}
\end{align}

the far-field diffraction intensity becomes

\begin{equation}
    I = \left|s_1 \Psi_a + s_2 \Psi_b\right|^2
    = s_1^2 \left|\Psi_a\right|^2
    + 2\, s_1 s_2 \operatorname{Re}\left[\bar{\Psi}_a \Psi_b\right]
    + s_2^2 \left|\Psi_b\right|^2.
    \label{eq:intensity_quadratic}
\end{equation}

Crucially, the intensity is a quadratic function of $(s_1, s_2)$ with coefficients determined entirely by the network's normalized predictions and the known probe. The Fourier-domain components $\Psi_a$ and $\Psi_b$ depend only on $(\tilde{a}_o, \tilde{b}_o)$ and $(a_p, b_p)$. This factorization is less tractable with an amplitude/phase decoder output: the exit wave expansion no longer separates into independent terms linear in the scaling parameters, yielding an objective with complex-valued coupling terms that complicates the numerical optimization. Note that this approach can be further extended to modeling multiple incoherent probe modes (see SI).

The scaling parameters admit a physical interpretation in terms of amplitude and phase contrast:
\begin{align}
    c_A &= \sqrt{s_1^2 + s_2^2}, &
    c_\phi &= \arctan\left(s_2 / s_1\right),
    \label{eq:contrast}
\end{align}
where $c_A$ characterizes the overall amplitude scaling of the reconstructed object relative to the normalized prediction, and $c_\phi$ captures the effective phase offset introduced by the independent scaling of real and imaginary components.

The real-imaginary representation has several key advantages over amplitude-phase: (1) Rectangular coordinates are linear through the Fourier transform, ensuring separability of the scaling constants in the final intensity expression; (2) Phase continuities at the $-\pi, \pi$ boundary can cause gradient problems and poor reconstruction fidelity. In rectangular coordinates, an arbitrary number of rotations can be performed without issue. This can result in the model learning an arbitrary phase rotation, but this does not affect the validity of the solution (further discussion in SI). (3) In rectangular coordinates, both branches are symmetric, allowing them to learn equally rich representations of the object. This is not the case in phase-amplitude, as the branches are not linearly coupled, sometimes forcing the neural network to learn the majority of sample variation in one branch.

\subsection{Scaling parameter optimization}
\label{sec:least_squares}

During training, $s_1$ and $s_2$ are treated as differentiable parameters and optimized jointly with the network weights by predicting the intensity using equation \ref{eq:intensity_quadratic}, and then backpropagating through the Poisson negative log-likelihood loss (see Section \ref{sec:PtychoPINN-overview}. At inference, the network weights are frozen and $(s_1, s_2)$ are determined by solving the least-squares formulation, equation \ref{eq:ls_objective}, after accumulating intensities across all inputs, requiring only the computation of $\Psi_a(\theta,P)$ and $\Psi_b(\theta,P)$ from the network predictions and the known probe.

Given a set of $N$ measured diffraction patterns $\{I_n^{\text{meas}}\}_{n=1}^{N}$ and the corresponding network predictions $\{(\tilde{a}_{o,n}, \tilde{b}_{o,n})\}_{n=1}^{N}$, the optimal scaling parameters are determined by minimizing the sum of squared residuals,
\begin{equation}
    \min_{s_1, s_2} \sum_{n=1}^{N} \sum_{\mathbf{q}} \left[
        I_n^{\text{meas}}(\mathbf{q})
        - I_n^{\text{pred}}(\mathbf{q};\, s_1, s_2)
    \right]^2,
    \label{eq:ls_objective}
\end{equation}

where $\mathbf{q}$ indexes detector pixels and $I_n^{\text{pred}}$ is given by Eq.~\eqref{eq:intensity_quadratic}. The scaling parameters can be calculated on a per-input, per-batch or per-dataset basis. In practice, we found that calculating $s_1$ and $s_2$ on a per-dataset basis, i.e. a two global parameters for a single dataset, still allows the network to focus on textures while preserving global contrast across the sample.

\begin{figure}[htbp]
    \centering
    \includegraphics[width=0.999\textwidth]{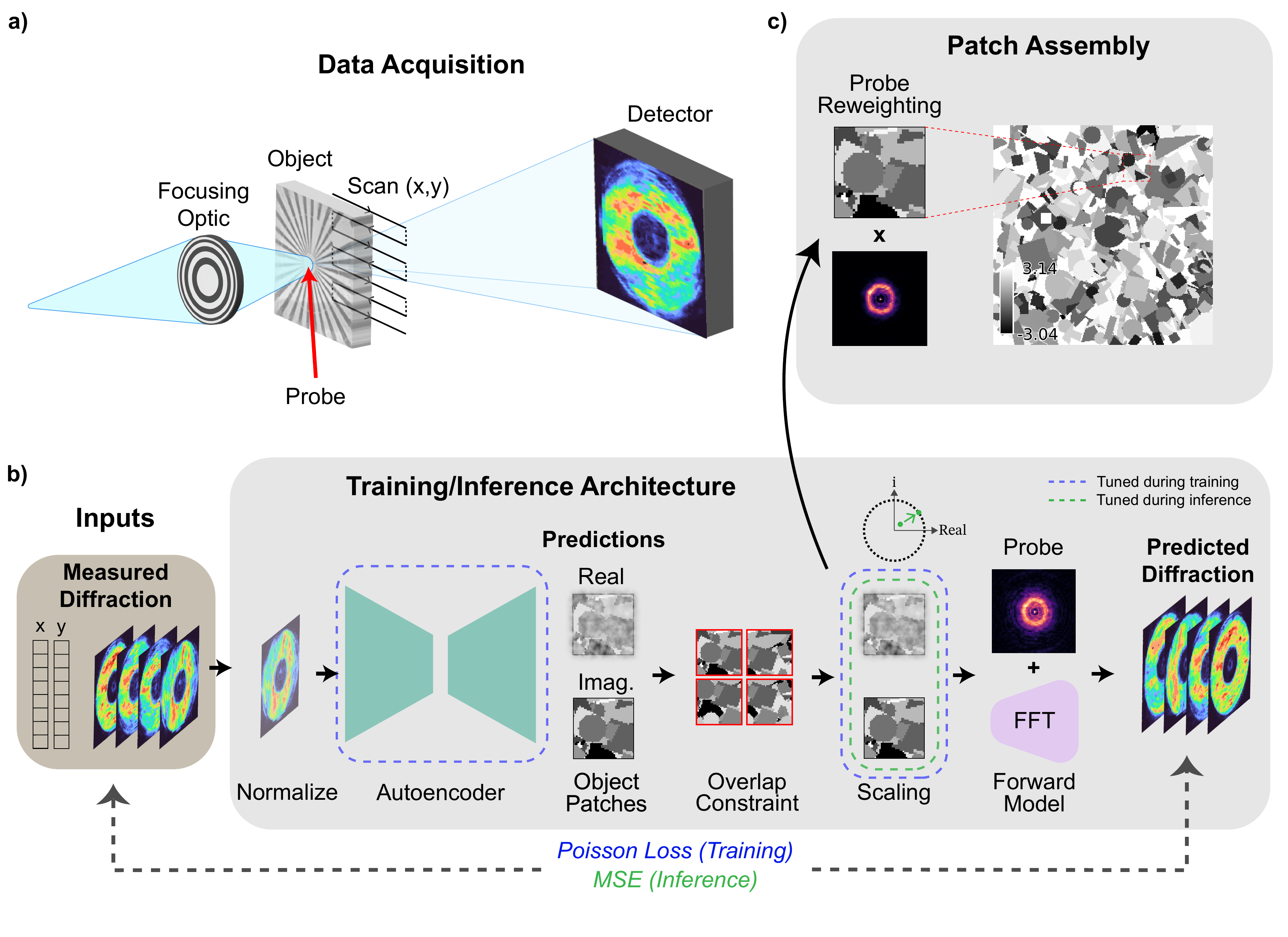}
    \caption{Overview of PtychoPINN-CI training and inference. (a) Overlapping diffraction patterns and their scan positions are collected during data acquisition; position information is enforced through the overlap consistency loss. (b) Normalized diffraction batches are passed through a U-net encoder-decoder, which outputs unit-less real and imaginary object predictions. Scaling parameters $(s_1, s_2)$ are fit by maximizing the Poisson likelihood between the forward-modeled exit wave and measured intensities. The same fitting procedure is applied at inference, producing measurement-consistent predictions on unseen datasets. (c) Each diffraction pattern is mapped to a local object patch, which is reweighted by the probe intensity $|P|^2$ before translation and averaging to form the full-field reconstruction.}
    \label{fig:model_overview}
\end{figure}

During training, the probe is normalized to improve gradient flow across the network. The probe is not normalized at inference, ensuring the scaled outputs are consistent with conventional reconstructions and the measurement itself (i.e. amplitudes around 1). This post-hoc scaling correction requires no retraining and adds negligible computational cost to the inference pipeline. An overview of the training pipeline and inference strategy can be found in Figure \ref{fig:model_overview}.

\subsection{Probe-weighted patch stitching}
\label{sec:probe_weighted_stitching}

In ptychography, the probe intensity $|P(\mathbf{r})|^2$ reflects the local measurement information: regions illuminated more strongly contribute more signal to the diffraction pattern and are reconstructed more reliably. We exploit this by assembling the full-field object from overlapping patches using probe-intensity weights,

\begin{equation}
    O_{\text{stitched}}(\mathbf{r})
    = \frac{\sum_i \left|P(\mathbf{r} - \mathbf{r}_i)\right|^2 O_i(\mathbf{r})}
           {\sum_i \left|P(\mathbf{r} - \mathbf{r}_i)\right|^2},
    \label{eq:probe_weighted_stitching}
\end{equation}

where $O_i$ is the reconstructed patch at scan position $\mathbf{r}_i$. This weighting naturally reduces contributions from the probe periphery, where measurement signal-to-noise is lowest, and suppresses boundary artifacts that arise from uniform averaging, which simply divides each pixel by the number of contributing images. The formulation mirrors the denominator normalization used in iterative ptychography algorithms~\cite{maiden2009improved, thibault2008high}, extending this established principle to DNN-based reconstruction. Probe-weighted stitching is applied during both training (for the overlap constraint evaluation) and inference for PtychoPINN-CI. This differs from most DNN methods, which to our knowledge use uniform averaging for image assembly.

\subsection{PtychoPINN: An unsupervised ptychography neural network}
\label{sec:PtychoPINN-overview}

PtychoPINN is an unsupervised convolutional neural network for ptychographic reconstruction~\cite{vong2025generalizable, hoidn2023physics}. The network accepts groups of overlapping diffraction patterns arranged with fixed relative coordinates and outputs complex-valued object patches (see original paper for details). A physics-based forward model, comprising of probe multiplication, Fourier transform, and intensity re-scaling, maps predicted objects back to diffraction space, enabling unsupervised training via a Poisson negative log-likelihood loss comparing predicted intensities with measurement intensities~\cite{hoidn2023physics}. Correctly scaling the predicted intensity is particularly important for this loss: the Poisson distribution's mean and variance are identical, so incorrectly scaled predictions distort both the expected photon count and its assumed noise level simultaneously, biasing gradient updates and degrading reconstruction fidelity. The loss is defined as:

\begin{equation}
    \mathcal{L}_{\text{Poiss}}
    \;=\;
    \sum_{n=1}^{N} \left[I_n^{\text{pred}}(\mathbf{n};\, s_1, s_2) -
        I_n^{\text{meas}}(\mathbf{n})
        \log (I_n^{\text{pred}})
    \right]^2,
    \label{eq:ls_poisson_objective}
\end{equation}

Real-space overlap constraints enforce consistency between adjacent predictions, avoiding reconstruction phase ambiguity~\cite{hoidn2023physics, vong2025generalizable}. The previous iteration of PtychoPINN, named PtychoPINN-torch, is based on an autoencoder design using additional Convolutional Block Attention Modules (CBAM) in the encoder. In PtychoPINN-CI, we have additionally added skip connections between encoder and decoder to form a U-net type autoencoder.

Raw diffraction image inputs to PtychoPINN are scaled using an amplitude normalization strategy for the kth diffraction pattern:

\begin{align}
  I_k \;=\; I'_k \cdot \sqrt{\frac{(N/2)^2}{\big\langle\sum_{i,j} |I'_{ij}|^2\big\rangle}},
  \label{eq:norm}
\end{align}

where I' represents the raw diffraction pattern, N is the image dimension (e.g. 64 pixels) and i,j represent image coordinates. This input scaling strategy, which we term RMS scaling, allows the neural network to be photon-scale invariant on the input side.

Supervised networks typically do not scale their outputs since they directly predict amplitude and phase. Unsupervised networks, on the other hand, require rescaling the normalized outputs to intensity photon units when reconstructing the input signal. PtychoPINN-torch uses an output scaling strategy that learns during training but does not change at inference, sometimes resulting in reconstruction artifacts such as extremely large amplitudes (i.e. much greater than 1) when predicting out-of-distribution.

PtychoPINN-torch also demonstrated that neural networks trained on synthetic diffraction data could generalize to experiment, showing high consistency in Fourier Ring Correlation versus models trained exclusively on experiment data. However, the authors noted a consistent bias in the predicted phase distributions relative to iteratively reconstructed references, with reduced phase contrast and haloing artifacts that suggested the synthetic training data did not capture the full complex-value structure of experimental objects~\cite{vong2025generalizable}.

\subsection{Real/imaginary synthetic object sampling}
\label{sec:real_imag_sampling}

In PtychoPINN-torch, synthetic training objects were generated by independently sampling amplitude values in $[0.5, 1.0]$ and phase values in $[-\pi, \pi]$, then converting to complex form~\cite{vong2025generalizable}. Reconstructing experimental data produces qualitatively consistent amplitude and phase images with higher spatial resolution than models exclusively trained on experiment data; however, the phase prediction also showed reduced phase contrast and haloing effects, which were absent in experiment-trained models.

We address this with several modifications to our synthetic object generation workflow: (1) Joint sampling of synthetic object components in rectangular coordinates instead of independent amplitude-phase sampling, (2) Ensuring high local phase contrast density, so that each input window contains numerous regions with strongly dissimilar phase values. (3) Incorporating empirical real-imaginary correlations that reflect the correlated material properties observed in experimental objects, versus independent uniform sampling in both dimensions.

For points (1) and (3), experimentally-measured objects are highly correlated in their real and imaginary values, due to their underlying material properties. These correlations need to be captured in the synthetic data, or the model will not learn the correct phase contrast scales on new test sets. This approach avoids the uniform sampling step that creates the distribution mismatch: instead of forcing synthetic objects to span the majority of the complex unit circle, the complex value distributions emerge naturally from correlations observed in real experimental conditions.

We explore several methods of correlation sampling, such as empirical correlations and Gaussian mixture models. In the simplest case, we construct an empirical distribution by normalizing a 2D histogram of complex values from a conventionally reconstructed object. We then sample synthetic object textures using this empirical distribution, resulting in complex value distributions that mirror the correlations observed in experiment data.

Maintaining high local phase contrast density is also critical. The global complex-value statistics can fit empirically-observed correlations, but the neural network only processes features at a local scale determined by the input window. This requires a sufficient number of regions within a single window that exhibit strong phase dissimilarity, which forces the neural network to differentiate highly contrasted components during a forward pass. Independent random sampling produces too many training samples with smooth phase gradients, which can limit the models' dynamic range.

% TODO: Specify the exact sampling ranges used for real/imaginary components.

% -----------------------------------------------------------------------------
\section{Results}
\label{sec:results}

\subsection{Experimental setup and baselines}
\label{sec:recon_overview}

We evaluate the proposed improvements using the same experimental datasets as the previous PtychoPINN-torch study~\cite{vong2025generalizable}, enabling direct comparison of reconstruction quality. These datasets span multiple instruments and facilities, including the Velociprobe (APS), Hard X-ray Nanoprobe (APS-CNM), Cosmic Imaging (ALS), and % TODO: FILL IN ALS-NS instrument name
; see the previous work for complete dataset descriptions~\cite{vong2025generalizable}. The baseline is the previous PtychoPINN-torch model with amplitude/phase decoder output, uniform patch stitching, and amplitude/phase synthetic object sampling. All models are trained using the training protocol described previously, and model comparisons use the same training and evaluation sets~\cite{vong2025generalizable}.

We compare five configurations that successively add components to demonstrate their compounding benefits:
\begin{enumerate}
    \item \textbf{Baseline}: amplitude/phase output without scaling (PtychoPINN-torch)
    \item \textbf{Real/imaginary output}: replaces the decoder with real/imaginary branches and adds dynamic test-time scaling
    \item \textbf{Probe-weighted stitching}: adds $|P|^2$-weighted patch reassembly to (2)
    \item \textbf{Single decoder}: merges the two decoder branches into a single branch that jointly predicts real and imaginary components
    \item \textbf{Synthetic data with correlated sampling}: trains exclusively on synthetic data generated using experimentally measured probe functions
\end{enumerate}

\begin{figure}[htbp]
    \centering
    \includegraphics[width=0.999\textwidth]{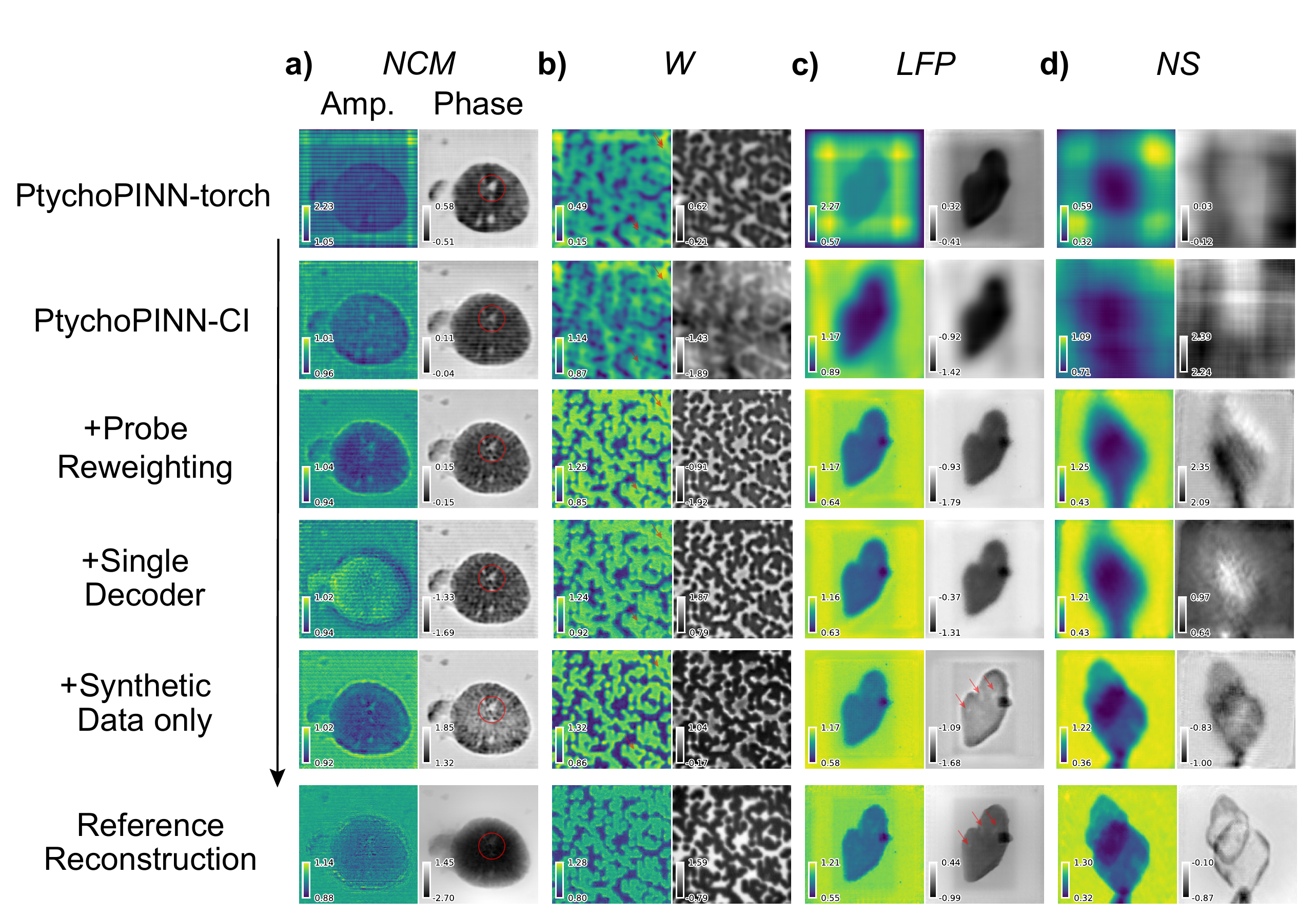}
    \caption{Example reconstructions on 4 distinct datasets from five model configurations, which incorporate additional components going downwards. A reference object at the top is generated using iterative reconstruction methods. Perceptual quality and resolution generally increases going downwards, as more components are added to the base model. The first four models were trained on exclusively experiment data, while the "Synthetic Data only" model was only trained on synthetic data. (a) NCM dataset(Velociprobe, Advanced Photon Source), consists of a centered nanoparticle with microscopic inner features and high contrast background. Models were trained only on the TP2 dataset measured on the same instrument. (b) W dataset(Hard X-ray Nanoprobe, APS), a high contrast tungsten test pattern. The amplitude predictions generally increase in resolution going downwards. (c) LFP dataset(Cosmic Imaging, Advanced Light Source), a catalytic nanoparticle. This dataset notably has less X-ray absorption, requiring non-trivial amplitude predictions which are successfully performed by the last 3 models. (d) NS (Cosmic Imaging, Advanced Light Source), overlapping nanoparticles. This dataset is challenging to reconstruct by training solely on experiment data, due to small dataset size. Combining model improvements with synthetic data leads to reasonable reconstruction quality.}
    \label{fig:recon_overview}
\end{figure}

Figure \ref{fig:recon_overview} shows reconstructions from all five configurations across four datasets with differing sample and measurement characteristics: (1) \textit{NCM} (Velociprobe), a LiNiCoMn$O_{2}$ nanoparticle with fine textured features; (2) \textit{W} (Hard X-ray Nanoprobe), a high-contrast tungsten test pattern; (3) \textit{LFP} (Cosmic Imaging), a catalyst nanoparticle with higher X-ray absorption and softer contrast; and (4) \textit{NS} % TODO: (instrument), description — FILL IN AFTER STEVE GIVES DETAILS
. The first four configurations are trained exclusively on experimental data, while the fifth uses only synthetic data generated from experimentally measured probes (see~\cite{vong2025generalizable} for details). Reference reconstructions (top row) are produced using the least-squares maximum likelihood algorithm implemented in ptychography software package Pty-Chi \cite{du2025pty}. Sample details can be found in Table \ref{tab:datasets} alongside fitted scaling constants.

\begin{table}[htbp]
    \centering
    \vspace{12pt}
    \begin{tabularx}{\textwidth}{@{}ccccX@{}}
    \toprule
    \textbf{Name} & \textbf{Source} & \textbf{\#} & \textbf{\shortstack{Photons\\per img.}} &
    \textbf{Description} \\
    \midrule
    \textit{FLY1} & Velociprobe & 10,304 & $4.4 \times 10^{5}$ & Pattern w/ background features\\
    \textit{NCM} & Velociprobe & 2,466 & $2.5 \times 10^{6}$ & LiNiCoMn$O_{2}$ particles \\
    \textit{W} & HXN & 25,921 & $9.6 \times 10^{6}$ & Tungsten test pattern\cite{babu2023deep} \\
    \textit{LFP} & Cosmic & 5,625 & N/A & Catalyst particle\cite{marcus2021ptychography}; acquired on CCD detector \\
    \textit{NS} & Cosmic & 960 & N/A & LiFeP$O_{4}$ particles \\
    \bottomrule
    \end{tabularx}
    \caption{Dataset details for all experiments used. All Velociprobe datasets were acquired at APS; HXN at APS-CNM; Cosmic at ALS.}
    \label{tab:datasets}
  \end{table}

Perceptual quality and resolution improve progressively from the PtychoPINN-torch baseline to the full PtychoPINN-CI configuration. PtychoPINN-torch either overestimates or underestimates amplitude, often producing unphysical values exceeding 1. Switching to the real/imaginary decoder with dynamic scaling immediately corrects these magnitudes, bringing amplitude predictions into closer agreement with the reference. With all components combined and synthetic data training (bottom row), PtychoPINN-CI reproduces fine-scale features present in the reference reconstructions (see indicators in Fig.~\ref{fig:recon_overview}), indicating the network provides higher feature resolution than PtychoPINN-torch.

PtychoPINN-CI also performs well on datasets with higher absorption. For \textit{LFP}, measured at lower X-ray energy, switching to the real/imaginary decoder with dynamic scaling alone does not substantially improve reconstruction quality relative to the baseline. A marked improvement occurs upon adding probe-weighted stitching, which we attribute to the extremely small probe used in this measurement: the probe intensity is concentrated in a small central region, meaning that much of each object patch prediction corresponds to weakly illuminated areas with low signal-to-noise. Probe reweighting downweights these noisy peripheral regions, significantly improving the assembled reconstruction. With all components and synthetic data training, subtle textured features visible in the reference are correctly reproduced (Figure \ref{fig:recon_overview}, red arrows), consistent with the improved reconstruction capacity from synthetic training observed previously~\cite{vong2025generalizable}. We note that training the baseline model on synthetic data does not achieve comparable quality \cite{vong2025generalizable}. \textit{NS} exhibits a similar pattern: the small probe and limited dataset size lead to poor reconstruction quality for the first two configurations, with noticeable improvement upon adding probe-weighted stitching, though resolution remains limited. Only the full PtychoPINN-CI configuration with synthetic data training yields adequate amplitude quality, though phase resolution is still reduced compared to the reference. Overall, these results demonstrate that this framework is generalizable, produces consistent reconstructions, and that all proposed components (real/imaginary output, dynamic scaling, probe reweighting, and correlated synthetic data training) contribute to the final reconstruction quality.

% TODO: Add summary table or figure of all configurations

\subsection{Ablation: real/imaginary vs.\ amplitude/phase output}
\label{sec:real_im_vs_amp_phase}

\begin{figure}[htbp]
    \centering
    \includegraphics[width=0.999\textwidth]{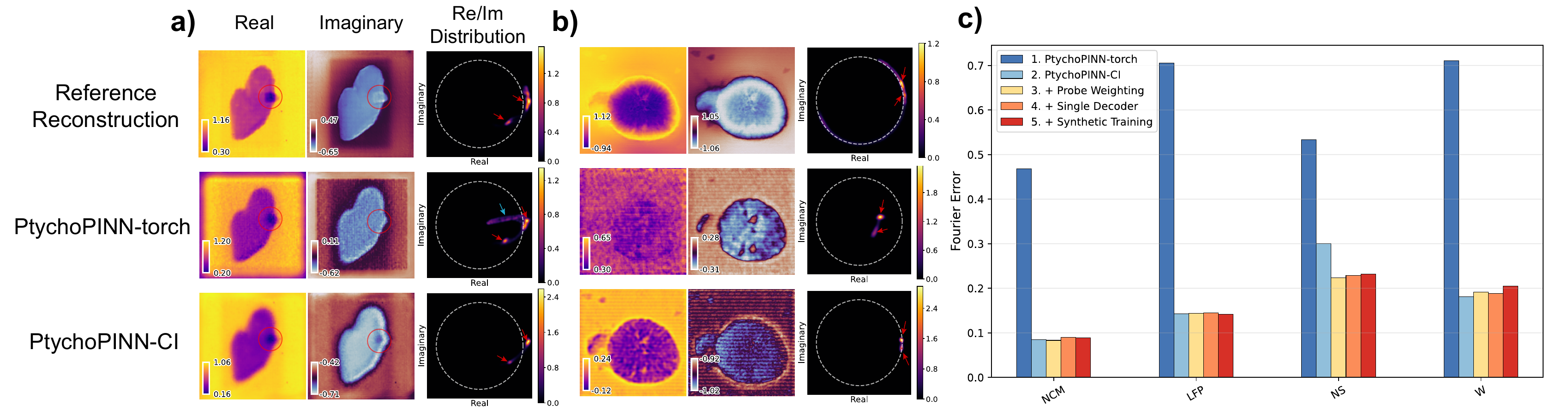}
    \caption{Real-imaginary representations of the LFP and NCM datasets for baseline PtychoPINN-torch and PtychoPINN-CI, both sharing the same training and evaluation datasets (a) The LFP reconstruction shows the PtychoPINN-torch prediction prioritizes learning the real object, while PtychoPINN-CI is more balanced, learning both channels equally. (b) The NCM reconstruction shows similar characteristics, where PtychoPINN-torch exclusively learns the imaginary features while ignoring any real components. (c) The change in Fourier error when comparing the baseline model with no scaling, to all PtychoPINN-CI which have dynamic test-time scaling applied. A large reduction in error shows the importance of test time scaling to produce physically-consistent predictions}
    \label{fig:ap_ri_comparison}
\end{figure}

We isolate the effect of the decoder representation by comparing two models that share inference-time scaling and probe-weighted stitching but differ only in their output parameterization. The Amplitude-Phase model adds these inference components to the PtychoPINN-torch decoder, while the Real-Imaginary model uses PtychoPINN-CI with separate real and imaginary decoder branches. Both models are trained on the same experimental datasets. Example reconstructions for \textit{LFP} and \textit{NCM} are shown in Figure \ref{fig:ap_ri_comparison}, plotted in real-imaginary coordinates to emphasize differences. An additional column shows the distribution of complex pixel values overlaid on the unit circle for each reconstruction.

Adding inference-time scaling to the amplitude-phase (AP) model produces measurement-consistent photon magnitudes, correcting the unphysical amplitudes seen in Fig.~\ref{fig:recon_overview}. Despite this improvement, the AP network prioritizes phase variation over amplitude during training, resulting in low-fidelity amplitude images. This imbalance is evident in the real-imaginary view (Fig.~\ref{fig:ap_ri_comparison}a): the AP model concentrates most information in one channel in both shown datasets. Inference-time scaling adjusts only the relative magnitudes of the real and imaginary components but does not redistribute learned information between channels; if the network underutilizes one channel during training, scaling alone cannot recover the unused model capacity. In contrast, the real-imaginary (RI) model distributes information more evenly across real and imaginary, which results in higher quality amplitude images (Fig. \ref{fig:recon_overview}) despite not being explicitly trained on amplitude. For example, the ``node''-like feature in \textit{LFP} is absent from the AP model's imaginary prediction but appears in both RI channels (red circles in Fig.~\ref{fig:ap_ri_comparison}a). For the NCM dataset (Fig. \ref{fig:ap_ri_comparison}b), the RI network still produces informative edge features in the imaginary representation despite the presence of additional noise, while the AP network's real prediction lacks any contrast.

Both models exhibit some degree of information distortion in their complex pixel value distributions (Fig.~\ref{fig:ap_ri_comparison}a, b, right columns). The most informative pixel modes (red arrows) are captured by both models, but are accompanied by spurious noise (AP model on \textit{LFP}, blue arrow) or a compressed dynamic phase range (both RI predictions). This is expected, as the autoencoder architecture compresses information, leading the model to prioritize the most informative sample features at the cost of encoding full dynamic phase contrast.

% For the NCM dataset (Fig.~\ref{fig:ap_ri_comparison}b), the AP model amplitude prediction has numerous low-resolution artifacts and has non-physical scaling above 1. In contrast, the real/imaginary model with scaling produces fundamentally different reconstructions: the factored representation allows the scaling optimization to correct the photon-scale mismatch independently of the object texture predicted by the network, resulting in physically-plausible amplitude scaling.

% We further note that the difference in amplitude reconstruction is fundamental the RI representation. An AP network learns to prioritize phase variation over amplitude, resulting in high quality phase predictions at the cost of low-fidelity and physically inaccurate amplitude images. On the other hand, due to equal contribution of real and imaginary predictions to form the amplitude image, an RI network has to place equal importance to both channels, resulting in more information throughput and a higher quality image (need to reword this).

We additionally compare the Fourier error across all model configurations from Fig.~\ref{fig:recon_overview}. The Fourier error quantifies the agreement between the predicted diffraction pattern produced by the neural network output and the raw measured data (see SI for details). PtychoPINN-torch(Fig.~\ref{fig:ap_ri_comparison}c, blue), which lacks test-time scaling, exhibits large Fourier errors and cannot correct its mismatched predictions. In contrast, all PtychoPINN-CI variants show a substantial reduction in Fourier error, consistent with the ability of dynamic scaling to match arbitrary measurement scaling at test time.

\subsection{Generating correlated synthetic data}
\label{sec:synthetic_data}

\begin{figure}[htbp]
    \centering
    \includegraphics[width=0.999\textwidth]{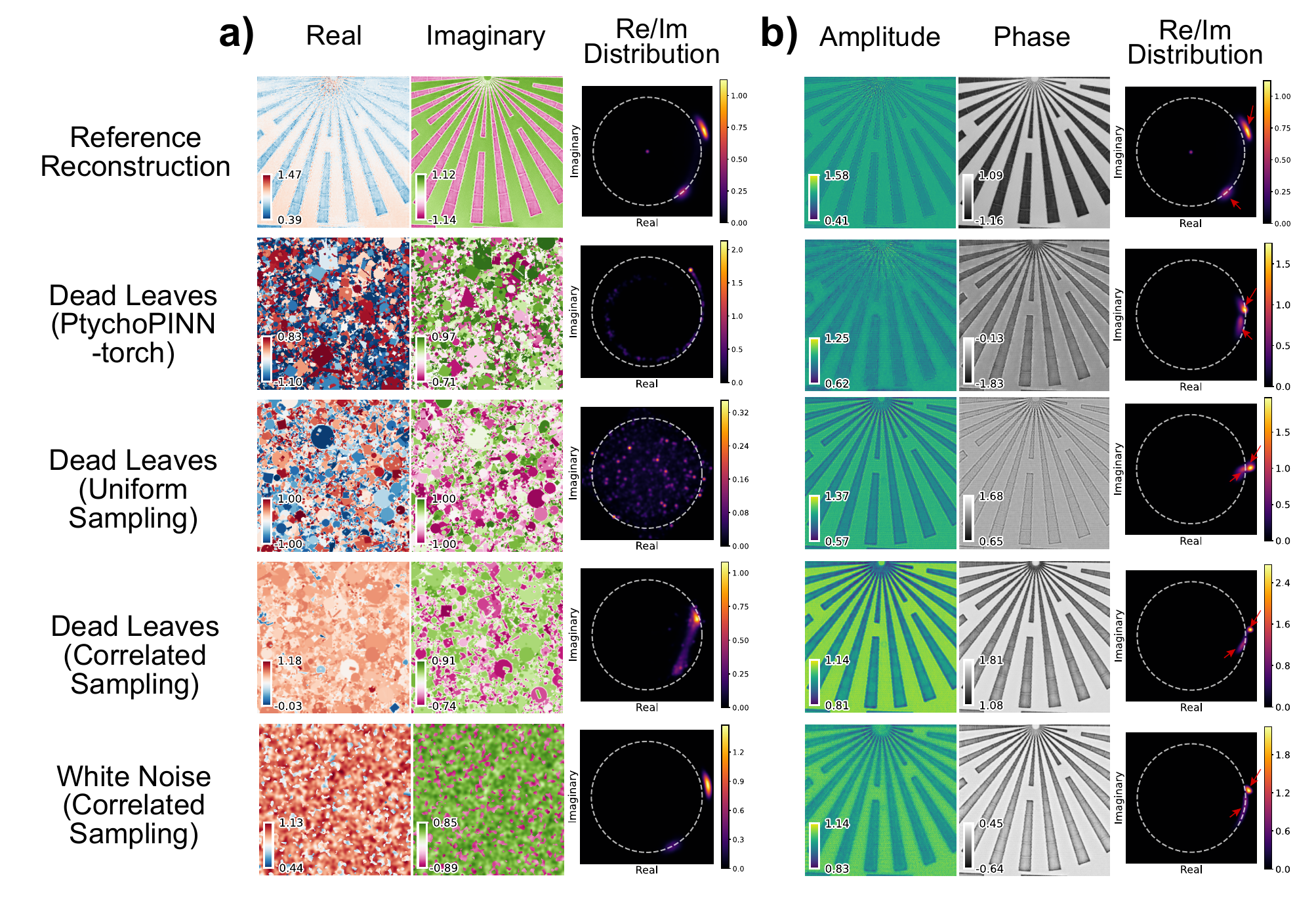}
    \caption{Synthetic dataset overview and corresponding predictions on the TP2 dataset. The evaluation dataset is unseen during model training. (a) A dataset overview, showing the reference reconstruction and synthetic datasets. From top to bottom: reference reconstruction of the TP2 Dataset, dead leaves from ref. \cite{vong2025generalizable}, dead leaves with uniform random real-imaginary sampling, dead leaves with correlated real-imaginary sampling from the reference object distribution, blurred white noise with the same correlation but high local phase contrast.  (b) Predicted objects from models trained on the corresponding per-row synthetic datasets. Background phase contrast is not properly captured by the first two dead leaves models, but is present in the last two models which were trained on more realistic amplitude-phase correlations}
    \label{fig:synthetic_data}
\end{figure}

In previous work, we demonstrated that synthetic datasets can train models that achieve reconstruction resolution comparable to or exceeding that of models trained exclusively on experimental data~\cite{vong2025generalizable}. These synthetic datasets combine experimentally measured probes with procedurally generated objects; for optimal reconstruction fidelity, the synthetic objects must span a broad range of spatial frequencies. While feature resolution generally improved with synthetic data training, we observed a systematic shift in the reconstructed phase and amplitude distributions, resulting in ``haloing'' artifacts at object edges and reduced background contrast. In this work, we address this issue through two contributions: first, our contrast-invariant network produces more accurate amplitude images through measurement-consistent scaling, as shown in Section \ref{sec:real_im_vs_amp_phase}. Second, we sample synthetic object values from an empirical real-imaginary distribution derived from reference reconstructions, thereby preserving correlations present in experimental data. Examples of synthetic objects are shown in Figure \ref{fig:synthetic_data}a alongside their real-imaginary distributions, which can be compared to the reference dataset \textit{TP2} (Figure \ref{fig:synthetic_data}a, top).

We compare four synthetic object classes to evaluate the effect of real-imaginary correlations on reconstruction quality. The first three are variants of Dead Leaves (DL), a procedural model of overlapping polygons with power-law-distributed radii~\cite{baradad2021learning}. All three share the same polygon radius distribution but differ in how real and imaginary values are assigned: (1) the original Dead Leaves implementation (DLO) derives correlated amplitude-phase values from randomly sampled refractive indices (see SI for details); (2) Dead Leaves Uniform (DLU) samples real and imaginary values independently and uniformly from $[-1, 1]$; and (3) Dead Leaves Correlated (DLC) draws real and imaginary values from the empirical joint distribution of the reference reconstruction (Figure \ref{fig:synthetic_data}a, top). The fourth class, White Noise Correlated (WNC), applies the same empirical sampling procedure as DLC but replaces the polygon geometry with low-pass-filtered white noise and incorporates high local phase contrast density. All correlated sampling distributions are derived from the \textit{TP2} dataset, which features spoke test patterns with high background contrast.

We train two model variants on the synthetic datasets shown in Figure \ref{fig:synthetic_data}a: PtychoPINN-torch on DLO, and PtychoPINN-CI independently on each of the remaining three datasets. Figure \ref{fig:synthetic_data}b shows the corresponding reconstructions. The model trained on DLO exhibits haloing at object edges and significantly reduced background phase contrast relative to the reference; its real-imaginary distribution reveals two distinct modes corresponding to test pattern and background, but with compressed phase angles consistent with the bias described in Section \ref{sec:real_im_vs_amp_phase}. These artifacts persist even when training PtychoPINN-CI on DLO, indicating that they originate from the synthetic data distribution rather than the network architecture (see SI). Training on DLU yields predictions with minimal phase contrast: without correlations between real and imaginary components, the model minimizes the average loss by collapsing the output distribution, as evidenced by the narrowly concentrated peaks in the real-imaginary scatter plot. In practice, experimentally measured objects exhibit strong real-imaginary correlations, and the training data must reflect this structure for the model to learn meaningful amplitude-phase relationships. Sampling from an empirical distribution, as in DLC and WNC (Figure \ref{fig:synthetic_data}b, bottom), preserves these correlations and yields reconstructions with background contrast and phase texture more consistent with the reference. Notably, DLC and WNC use fundamentally different spatial geometries (overlapping polygons versus filtered white noise) yet produce reconstructions that are difficult to distinguish quantitatively; understanding which synthetic object properties most influence reconstruction fidelity remains an open question for future work.

Sampling from a dataset-specific empirical distribution raises the concern of over-biasing the network toward a specific reference object, which could limit its generalization capabilities. However, we argue that the presence of realistic real-imaginary correlations in training data is more important than matching the specific correlations of the target dataset. We demonstrate this by predicting on all Velociprobe-measured datasets using a single model trained on DLC which sampled from the TP2 dataset. The \textit{NCM} reconstruction is shown in Fig.~\ref{fig:recon_overview}a, while other reconstructions can be found in the SI. In all cases, the phase contrast artifacts that accompanied previous synthetic data approaches are substantially reduced, which shows that the network remains robust even when the training dataset is partially biased. We note, however, that this generalization remains probe-specific: with a known probe, the model supports inference on objects with a wide range of admissible features and experiment conditions~\cite{vong2025generalizable}.

% TODO: Insert figure showing side-by-side amplitude/phase reconstructions
% TODO: Insert FRC-AUC comparison across datasets

% -----------------------------------------------------------------------------
\section{Discussion}
\label{sec:discussion}

The dynamic scaling factorization in PtychoPINN-CI transforms unit-less neural network outputs to match experiment data with minimal overhead. % TODO: Include quantitative overhead numbers
Combined with synthetic data training and realistic real-imaginary correlations, this workflow reduces the phase distortion observed in the previous generation of PtychoPINN while improving reconstruction resolution. However, PtychoPINN-CI still exhibits phase compression in its predictions despite improved phase contrast. We attribute this to two possible factors: (1) the 2.8 million parameter network may lack sufficient capacity to represent the full range of complex-value correlations present in experimental objects, or (2) the synthetic training data may still lack fundamental structure found in real measurements. Despite this limitation, PtychoPINN-CI with 30\% fewer parameters than PtychoPINN-torch achieves improved reconstruction quality across all evaluated datasets, demonstrating the effectiveness of combining an unsupervised learning approach with a principled scaling strategy. Part of the parameter reduction comes from merging the two separate real and imaginary decoder branches into a single branch. Because real and imaginary object components are correlated in practice, separate branches risk learning redundant representations; a single branch that jointly predicts both components eliminates this redundancy while achieving equivalent reconstruction quality (Fig.~\ref{fig:recon_overview}, configurations 3 vs.\ 4). We expect that incorporating the scaling framework into more expressive architectures, such as vision transformers, will address phase compression while retaining the benefits of dynamic scaling.

We envision two primary applications for PtychoPINN-CI. First, as identified in previous work~\cite{vong2025generalizable}, a trained DNN can rapidly reconstruct new measurements for experiment steering. The compact model size (2.8 million parameters) enables deployment in resource-constrained beamline environments, and the scaling factorization improves the quantitative reliability of these reconstructions, particularly for samples with higher absorption such as the \textit{LFP} dataset (Fig.~\ref{fig:recon_overview}c). Second, because the scaling factorization produces photon-scale-consistent outputs, these reconstructions could serve as initialization for conventional iterative algorithms, allowing them to bypass coarse object refinement and begin fine-scale iteration. The residual phase compression may need to be addressed before neural network predictions can fully replace early-stage iterative refinement, but the correctly scaled amplitude provides a meaningful starting point.

The deployment workflow for PtychoPINN-CI is straightforward: at the beginning of an experimental campaign, the neural network is trained on any combination of synthetic or experimental data from initial measurements. Synthetic data generation requires only a probe function, which can be obtained from previous measurements or from a conventional reconstruction taken during the running experiment. The trained network can then be applied to subsequent measurements and should remain robust to fluctuating photon scales provided the probe geometry does not substantially change.
% -----------------------------------------------------------------------------
\section{Conclusion}
\label{sec:conclusion}

We presented PtychoPINN-CI, a scaling framework for ptychographic neural networks that decouples learned object texture from dataset-dependent measurement scales. The central insight is that a real-imaginary decoder representation preserves linearity through the Fourier transform, allowing the far-field intensity to be factored as a quadratic function of two scaling parameters $(s_1, s_2)$. These parameters are efficiently optimized at test time, converting unit-less network predictions into measurement-consistent reconstructions without retraining. We demonstrated this approach across multiple instruments and facilities, showing that a single trained network produces correctly scaled amplitude and phase images despite order-of-magnitude differences in photon flux between datasets. The real-imaginary representation additionally balances information across both output channels, yielding higher fidelity amplitude reconstructions than the conventional amplitude-phase decoder, while probe-intensity-weighted stitching further improves image quality by incorporating spatially varying measurement confidence.

This scaling framework extends naturally to synthetic data training. By sampling synthetic object values from empirical real-imaginary distributions, we resolve the phase distribution mismatch identified in previous work~\cite{vong2025generalizable} and enable synthetic-only training with minimal amplitude-phase distortions. The deployment workflow requires only a probe function to generate synthetic training data, after which the trained network generalizes to subsequent measurements on the same instrument with measurement-consistent outputs suitable for experiment steering or as initialization for iterative algorithms. These contributions are architecture-agnostic and transfer directly to more expressive backbones without modification to the scaling or sampling strategy, lowering a key barrier to deploying neural network reconstruction in X-ray ptychography experiments.

% -----------------------------------------------------------------------------
\section*{Acknowledgements}
The authors thank Ming Du for continued development and support for the ptychography reconstruction software Pty-Chi.

\section*{Funding}
This work is supported by the U.S. Department of Energy (DOE) Office of Science-Basic Energy Sciences (BES) awards Collaborative Machine Learning Platform for Scientific Discovery and Collaborative Machine Learning Platform for Scientific Discovery 2.0. Additional support was provided by the U.S. DOE, Office of Science-Advanced Scientific Computing Research (ASCR) and BES award X-ray \& Neutron Scientific Center for Optimization, Prediction \& Experimentation (XSCOPE). This work also received support from the DOE Office of Science ASCR Leadership Computing Challenge (ALCC) through the 2025–2026 award Enhancing APS-Enabled Research through Integrated Research Infrastructure. This research used resources of the Advanced Photon Source (APS) and the Argonne Leadership Computing Facility (ALCF), both U.S. DOE Office of Science user facilities operated by Argonne National Laboratory under Contract No. DE-AC02-06CH11357. This research used resources of the Advanced Light Source (ALS), a U.S. DOE Office of Science user facility operated by Lawrence Berkeley National Laboratory under Contract No. DE-AC02-05CH11231. This research used resources of the Linac Coherent Light Source (LCLS), a U.S. DOE Office of Science user facility operated by SLAC National Accelerator Laboratory under Contract No. DE-AC02-76SF00515.

The U.S. Government retains for itself, and others acting on its behalf, a paid-
up nonexclusive, irrevocable worldwide license in said article to reproduce, prepare derivative works, distribute copies to the public, and perform publicly and display publicly, by or on behalf of the Government. The Department of Energy will provide public access to these results of federally sponsored research in accordance with the DOE Public Access Plan. http://energy.gov/downloads/doe-public-access-plan.

\section*{Disclosures}
The authors declare no conflicts of interest.

\section*{Data Availability}
Datasets and model artifacts shared with the original PtychoPINN-torch paper can be found at: \url{https://doi.org/10.5281/zenodo.16968020}. Some new datasets used in this paper are not available at this time but may be obtained from the authors upon reasonable request. The PtychoPINN source code is available at \url{https://github.com/hoidn/PtychoPINN}.

\clearpage

% Title for supplemental information
\begin{center}
{\large \textbf{SUPPLEMENTAL INFORMATION}}

\vspace{8pt}

{\normalsize \textbf{Contrast-invariant deep ptychography neural networks}}

\vspace{4pt}

{\small Albert Vong, Steven Henke, Oliver Hoidn, Hanna Ruth, Junjing Deng, Alexander Hexemer, Apurva Mehta, Nicholas Schwarz}
\end{center}

\vspace{12pt}

\section{Supplemental Information}

\subsection{Experimental dataset details}

Ptychodus is a software package from the Advanced Photon Source that acts as a ptychography data pipeline. It has integrated capabilities with the Pytorch-optimized iterative package Pty-Chi, which was used to reconstruct ground-truth data for experimental reconstruction verification.

All ground truth reference data in this manuscript was generated using the least-squares maximum likelihood algorithm at 5000 iterations, with a single probe mode and no position refinement. The object reconstruction (which we define as reference reconstruction in the main text), alongside pixel-standardized diffraction images and position data are packaged into individual .npz files (per experiment) for training and inference. Additional dataset details can be found in \cite{vong2025generalizable}.

\subsection{Incorporating multiple incoherent probe modes}

The formulation outlined in section 2.1 of the main text can easily be extended to multiple probe modes, i.e.
\begin{equation}
    % Global R-factor for multiple images
    I_{measured} = \sum_{i=1}^{N}|\mathcal{F}\left[P_i \cdot O\right]|^{2}
    \label{R-factor equation}
\end{equation},

where $I_{measured}$ represents the sum of exit wave contributions from N considered incoherent probe modes. This slightly modifies the final training objective, which now becomes

\begin{equation}
    I = \sum_{i=1}^{N}\left|s_a \Psi_{a,i} + s_b \Psi_{b,i}\right|^2
    = s_a^2 \sum_{i=1}^{N}\left|\Psi_{a,i}\right|^2
    + 2\, s_a s_b \sum_{i=1}^{N}\operatorname{Re}\left[\bar{\Psi}_{a,i} \Psi_{b,i}\right]
    + s_b^2 \sum_{i=1}^{N}\left|\Psi_{b,i}\right|^2.
    \label{eq:si_intensity_quadratic_multiprobe}
\end{equation}

since the scaling constants $s_1$ and $s_2$ can be factored out of the summed expressions as well. This means the multiple probe modes are only considered when calculating the final intensity through the physics forward model. It does not affect any other components of the neural network training workflow.m

\subsection{PtychoPINN-CI reconstructions}

\begin{figure}[htbp]
    \centering
    \includegraphics[width=0.999\textwidth]{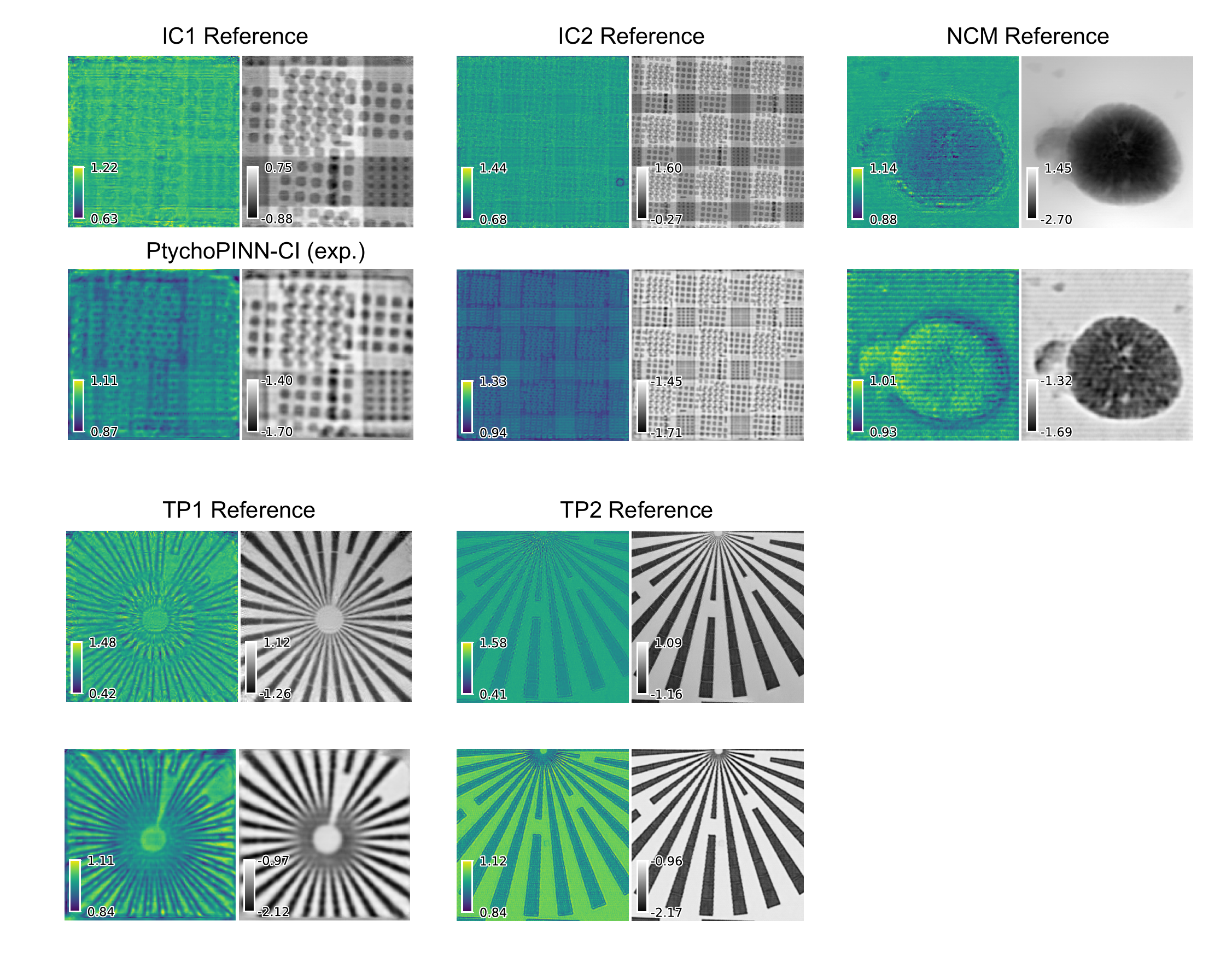}
    \caption{Reconstructions of experimental datasets measured on the Velociprobe using PtychoPINN-CI trained only on TP2. From left to right: IC1, IC2, NCM, TP1 and TP2}
    \label{fig:exp_results}
\end{figure}

\begin{figure}[htbp]
    \centering
    \includegraphics[width=0.999\textwidth]{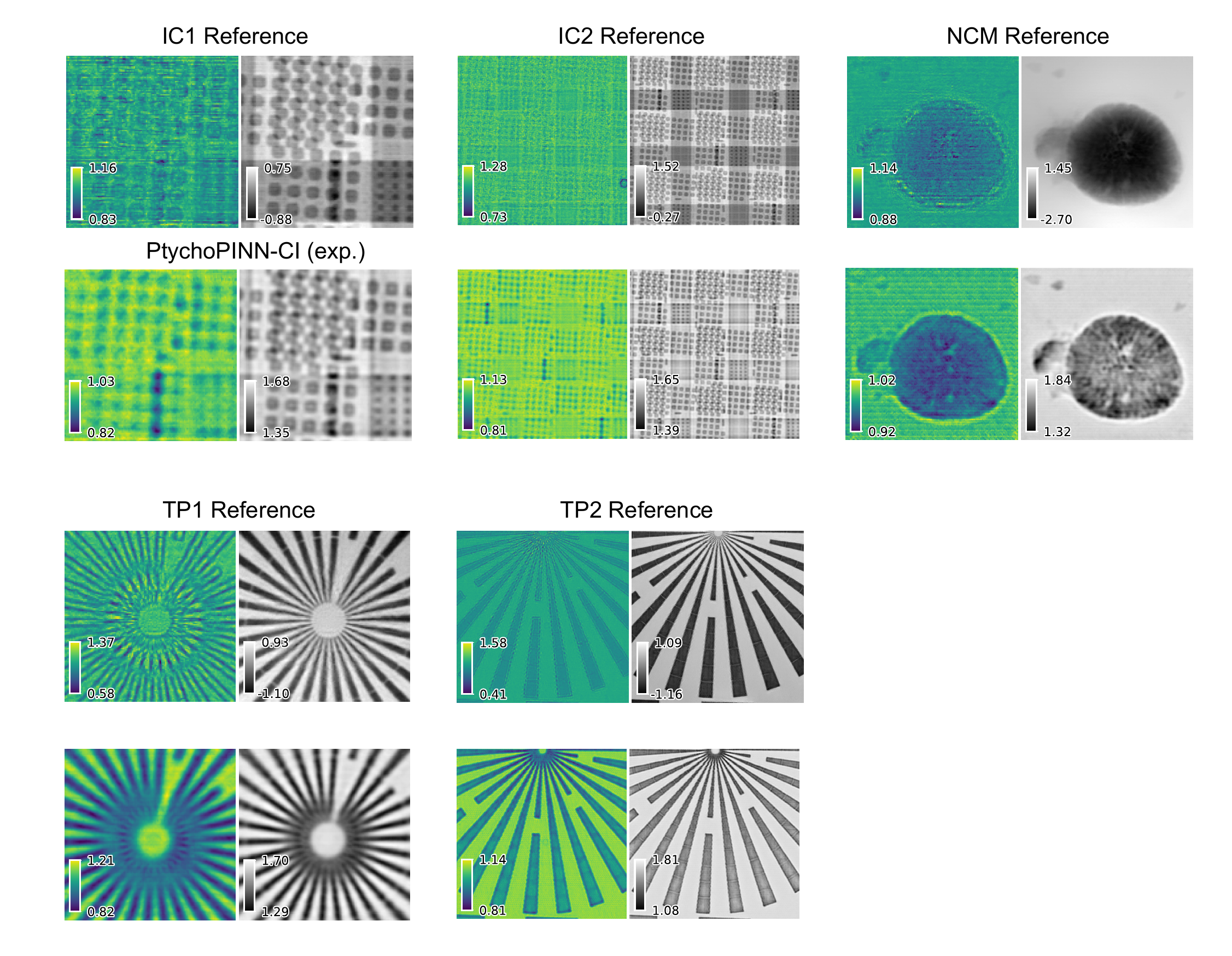}
    \caption{Reconstructions of experimental datasets measured on the Velociprobe using PtychoPINN-CI trained only on synthetic data derived from the TP2 reconstructed probe. From left to right: IC1, IC2, NCM, TP1 and TP2}
    \label{fig:syn_results}
\end{figure}

Figures S\ref{fig:exp_results} and S\ref{fig:syn_results} show all reconstructions measured on the Velociprobe instrument at the APS, using a model that was trained either directly on the TP2 dataset, or a model that was trained on synthetic data derived from the TP2 probe. We can observe good generalization from the TP2-trained model to other models measured on the Velociprobe, owing to the similarity between train and test probes. Interestingly, we observe better generalization from the experiment-based model to other datasets with PtychoPINN-CI, compared to our observations for PtychoPINN-torch \cite{vong2025generalizable}. We attribute this to more balanced channel learning, alongside enhanced resolution from probe-weighted stitching.

\subsection{Fourier error calculation}

Fourier error describes the difference between measured and predicted intensities across all pixels in a diffraction image. When normalized against the measured image itself, we get a modified Fourier error metric called the R-factor. We calculate the R-factor using the following formula for all datasets,

\begin{equation}
    % Global R-factor for multiple images
    R = \frac{\sum_{i=1}^{N} \sum_{u,v} \left| \sqrt{I_{meas,i}(u,v)} - \sqrt{I_{pred,i}(u,v)} \right|}{\sum_{i=1}^{N} \sum_{u,v} \sqrt{I_{meas,i}(u,v)}}
    \label{R-factor equation}
\end{equation}

where $(u, v)$ represent pixel coordinates while $i$ represents the image index within a given dataset. This version of the R factor averages across the entire dataset.

\subsection{Phase rotation offset}

When learning with PtychoPINN-CI in real and imaginary coordinates, the output distribution may not be phase-centered at $1 + 0i$, since a reference point is not explicitly enforced in the loss function nor learned by the network. We find in practice this does not affect the amplitude and phase images which are agnostic to the relative orientation of the predicted values. All predictions shown in the main text have been phase-centered for ease of comparison with the reference reconstructions which have been centered as a convention. A small loss penalization term can be added to phase-center the predicted object values around $1 + 0i$, but we find in practice this can sometimes affect PtychoPINN-CI's consistency at learning the correct representation.

\bibliographystyle{unsrtnat}
\bibliography{references}

\end{document}